\documentclass[letter]{spie}

\usepackage[utf8]{inputenc}
\usepackage{graphicx}
\usepackage{siunitx}
\usepackage{capt-of}
\usepackage{cleveref}
\usepackage{array}
\usepackage{caption}
\usepackage{subcaption}
\usepackage{amsmath,amssymb}

\usepackage{booktabs}       
\usepackage{makecell}
\usepackage{arydshln}       

\usepackage[colorlinks=true, allcolors=blue]{hyperref}

\title{Towards increased trustworthiness of deep learning segmentation methods on cardiac MRI}

\author{%
    Jörg Sander\supit{a}, Bob D. de Vos\supit{a}, Jelmer M. Wolterink\supit{a} and Ivana Išgum\supit{a}%
    \skiplinehalf%
    \small%
    \supit{a}Image Sciences Institute, University Medical Center Utrecht, Utrecht, The Netherlands\\%
}

\authorinfo{Send correspondence to J.Sander (email: j.sander@umcutrecht.nl)}

\begin{document}
    \maketitle 
    
    \begin{abstract}
    Current state-of-the-art deep learning segmentation methods have not yet made a broad entrance into the clinical setting in spite of high demand for such automatic methods. One important reason is the lack of reliability caused by models that fail unnoticed and often locally produce anatomically implausible results that medical experts would not make. This paper presents an automatic image segmentation method based on (Bayesian) dilated convolutional networks (DCNN) that generate segmentation masks and spatial uncertainty maps for the input image at hand. The method was trained and evaluated using segmentation of  the left ventricle (LV) cavity, right ventricle (RV) endocardium and myocardium (Myo) at end-diastole (ED) and end-systole (ES) in 100 cardiac 2D MR scans from the MICCAI 2017 Challenge (ACDC). Combining segmentations and uncertainty maps and employing a human-in-the-loop setting, we provide evidence that image areas indicated as highly uncertain, regarding the obtained segmentation, almost entirely cover regions of incorrect segmentations. The fused information can be harnessed to increase segmentation performance. Our results reveal that we can obtain valuable spatial uncertainty maps with low computational effort using DCNNs.
    \end{abstract}   
   
     
    \section{Purpose}
     
	Decisions by medical experts are increasingly enriched and augmented by intelligent machines e.g., through computer aided diagnosis (CAD). The quality of the joint decision process would improve if the automatic systems were able to indicate their uncertainty. This assumes that the provided uncertainty information is reliable i.e., valuable to be considered. A system indicating high uncertainty in image areas of incorrect segmentations could be used to detect and subsequently refer these regions to medical experts. Applying such a human-in-the-loop setting would result in increased segmentation performance. In addition, such a setting could mitigate a severe deficiency of current state-of-the-art deep learning segmentation methods which occasionally generate anatomically implausible segmentations\cite{bernard2018deep} that a medical expert would never make.
	Previous research has mainly focused on the assessment of uncertainty in disease prediction \cite{leibig2017leveraging}, tissue segmentation \cite{kwon2018uncertainty} and pulmonary nodule detection \cite{ozdemir2017} by utilizing Bayesian neural networks (BNN) or test-time data augmentation techniques \cite{ayhan2018test}. Additional methods to estimate uncertainty are Deep Ensembles \cite{lakshminarayanan2017simple} and Learned Confidence Estimates \cite{devries2018learning}. In the former multiple models are trained and the variance of their predictions is used as confidence measure, whereas in the latter the model outputs a confidence measure simultaneously with the prediction. 
	
	In this work, using multi-structures segmentation in cardiac MR images, we introduce a method that simultaneously generates segmentation masks and uncertainty maps by using a dilated convolutional network (DCNN). We compare two approaches to obtain uncertainty maps. First, we use entropy maps (e-maps) that can be efficiently generated by any probabilistic classifier as entropy is a theoretically grounded quantification of uncertainty in information theory. Second, we employ Bayesian uncertainty maps (u-maps) that are obtained by Bayesian DCNNs (B-DCNN).
	In addition, we reveal that a valuable uncertainty measure can be obtained if the applied model is \textit{well calibrated}, i.e. if generated probabilities represent the likelihood of being correct. We demonstrate this by simulating a human-in-the-loop setting and provide evidence that image areas indicated as highly uncertain regarding the obtained segmentation almost entirely cover regions of incorrect segmentations. Hence, the fused information can be employed in clinical practice to inform an expert whether and where the generated segmentation should be adjusted.

    \section{Data description}
      
     Data from the MICCAI  \num{2017} Challenge on automated cardiac diagnosis (ACDC) \cite{bernard2018deep} was used. The dataset consists of cardiac cine MR images (CMRI) from 150 patients who have been clinically diagnosed in five classes: normal, dilated cardiomyopathy, hypertrophic cardiomyopathy, heart failure with infarction, or right ventricular abnormality. Cases are uniformly distributed over classes. Manual reference segmentations of the left ventricle (LV) cavity, right ventricle (RV) endocardium and myocardium (Myo) at end-diastole (ED) and end-systole (ES) are provided for \num{100} cases. For each patient, short-axis (SA) CMRIs with 28-40 frames are available, in which the  ED and ES frame have been indicated. On average images consist of nine slices where each slice has a spatial resolution of \num{235}$\times$\num{263} voxels (on average). The image slices cover the LV from the base to the apex. In-plane voxel spacing varies from \num{1.37} to \SI{1.68}{\milli\meter}, with slice thickness from \num{5} to \SI{10}{\milli\meter} and sometimes inter-slice gap of \SI{5}{\milli\meter}. To correct for differences in voxel size, all 2D image slices were resampled to \num{1.4}$\times\SI{1.4}{\milli\meter}^2$. Furthermore, to correct for intensity differences among images, each MR volume was normalized between [\num{0.0}, \num{1.0}] according to the \num{5}th and \num{95}th percentile of intensities in the image. For detailed specifications about the acquisition protocol we refer the reader to Bernard et al. \cite{bernard2018deep}.
      
    \section{Method}
       
     To perform segmentation of tissue classes in cardiac 2D MR scans, we used the DCNN developed by Wolterink et al. \cite{wolterink2017automatic}. The DCNN architecture comprises a sequence of ten convolutional layers with increasing levels of kernel dilation which results in a receptive field for each voxel of \num{131}$\times$\num{131} voxels, or \num{18.3}$\times$ $\SI{18.3}{\centi\meter}^2$. The network has two input channels which take ED and ES slices as its input. We assume that the network leverages cardiac motion differences between ED and ES time points in order to better localize the target structures. To simultaneously segment the LV, RV, myocardium and background in ED and ES, the network has eight output channels where each channel provides a probability for one of the classes. Softmax probabilities are calculated over the four tissue classes for images acquired in ED and ES. To enhance generalization performance, the model uses batch normalization and weight decay.  
     
     To acquire spatial uncertainty maps of the segmentation during testing, two different approaches were evaluated. First, to obtain entropy maps (e-maps) we computed the multi-class entropy per voxel. Second, to obtain Bayesian uncertainty maps (u-maps), we implemented \textit{Monte Carlo dropout} (MC dropout) introduced by Gal \& Ghahramani \cite{gal2016dropout} for approximate Bayesian inference. We added dropout as the last operation in all but the final layer (by randomly switching off 10 percent of a layer's hidden units). By enabling dropout during testing, softmax probabilities are obtained with 10 samples per voxel. As an overall measure of uncertainty we used the maximum softmax variance per voxel over all classes\label{ref:maximum_variance}. The variance per voxel per class is obtained from the softmax samples for each class. We chose to use the maximum instead of the mean (as e.g., utilized by Leibig et al.\cite{leibig2017leveraging}) because we found that averaging attenuates the uncertainties. 
     
     The quality of e-maps and u-maps depends on the calibration of the acquired probabilities. Previous work\cite{lakshminarayanan2017simple} revealed that loss functions differ regarding how well the generated probabilities represent the likelihood of being correct. Therefore, we trained the model with three different loss functions: soft-Dice (SD), cross-entropy (CE), and the Brier score (BS) \cite{brier1950verification}, which is equal to the average gap between softmax probabilities and the references. This provides information about accuracy and uncertainty of the model. Computationally the Brier score loss is equal to the squared error between the one-hot encoding of the correct label and its associated probability.
     
    To use four-fold cross-validation we split the dataset into \num{75} and \num{25} training and test patients, respectively. Each model is evaluated on the holdout test images and we report combined results for all \num{100} patients. During training we used images with \num{151}$\times$\num{151} voxel samples, padded to  \num{281}$\times$\num{281} to accommodate the \num{131}$\times$\num{131} voxel receptive field. Training samples were augmented by \num{90} degree rotations of the images and references. The model was trained for 150,000 iterations using the snapshot ensemble technique described in \cite{huang2017snapshot}, while after every 10,000th iteration we reset the learning rate to its original value of \num{0.02} and stored the model. We used mini-batches of size \num{16} and applied Adam \cite{kingmadp} as stochastic gradient descent optimizer. To compare u-maps with e-maps at test time each model was evaluated twice. First, to obtain u-maps we used the last six stored models (iterations 100,000 to 150,000) of each fold to obtain segmentation results. Tissue class per voxel was determined using the mean softmax probabilities over 60 samples (10 samples per voxel per model). In addition, these probabilities served to compute the maximum variance (as described in the beginning of this section). Second, to obtain e-maps we solely employed the last stored model of each fold to acquire segmentation results. We disabled dropout during inference and used one forward pass to compute the softmax probabilities and determine the tissue class per voxel. The corresponding e-maps were computed as the entropy in the four-class probability distribution. Finally, for both evaluations as a post-processing step, the 3D probability volumes were filtered by selection of the largest \num{3}D \num{6}-connected component for each class. 
    
    The models were implemented using the PyTorch \cite{paszke2017automatic} framework and trained on one Nvidia GTX Titan X GPU with 12 GB memory.
    
	\section{Results and Discussion}
	\begin{figure}
			\begin{minipage}{0.40\textwidth}
				\begin{subfigure}{\textwidth}
				\centering
				\includegraphics[width=0.99\textwidth, trim=0 0 0 20mm, clip]{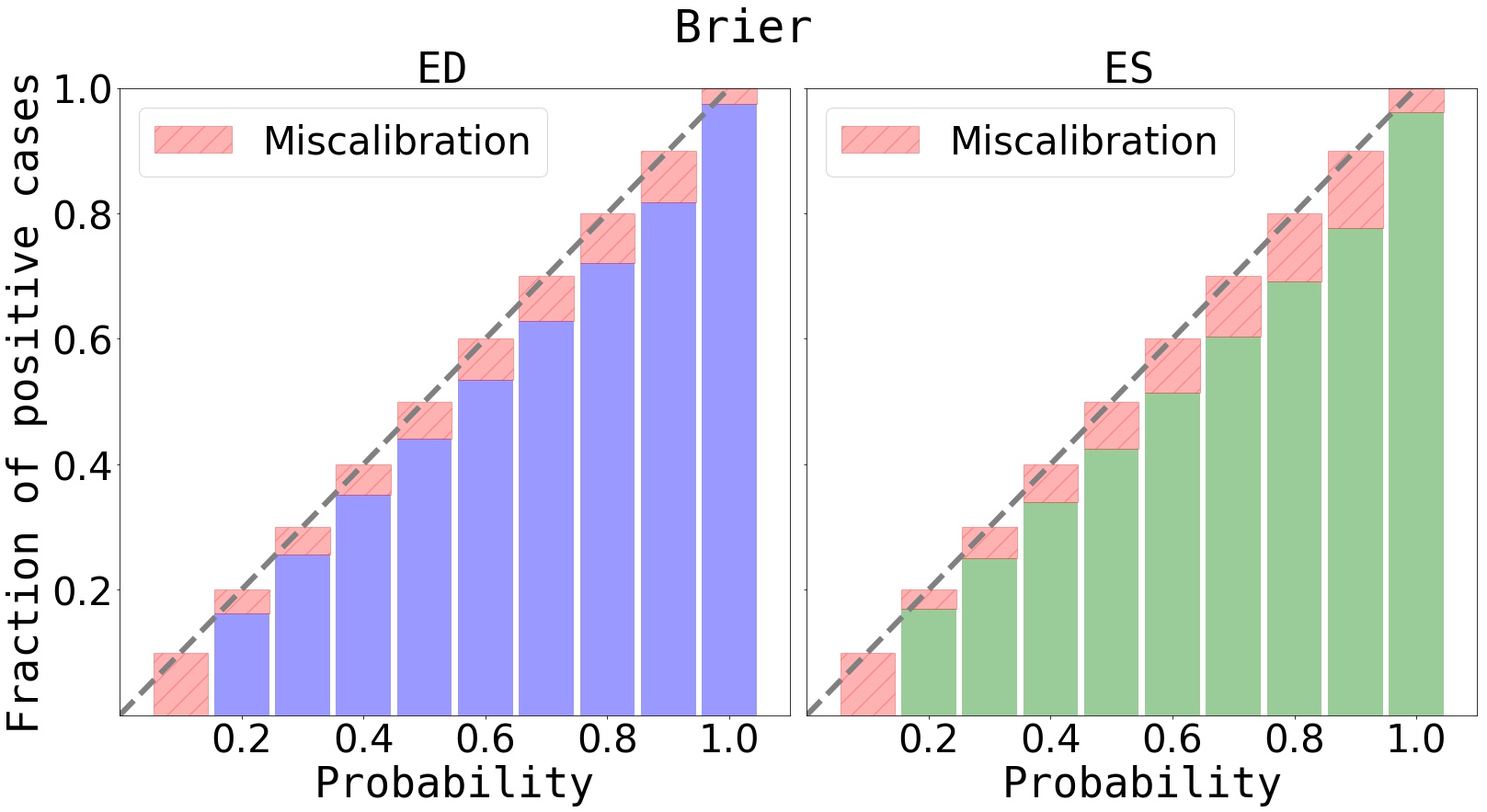}
				\caption{Brier score loss}
				\label{fig:calibration_brier}
				 \end{subfigure}%
			\end{minipage}%
			\centering
			\begin{minipage}{0.40\textwidth}
				\begin{subfigure}{\textwidth}
				\centering
				\includegraphics[width=0.99\textwidth, trim=0 0 0 20mm, clip]{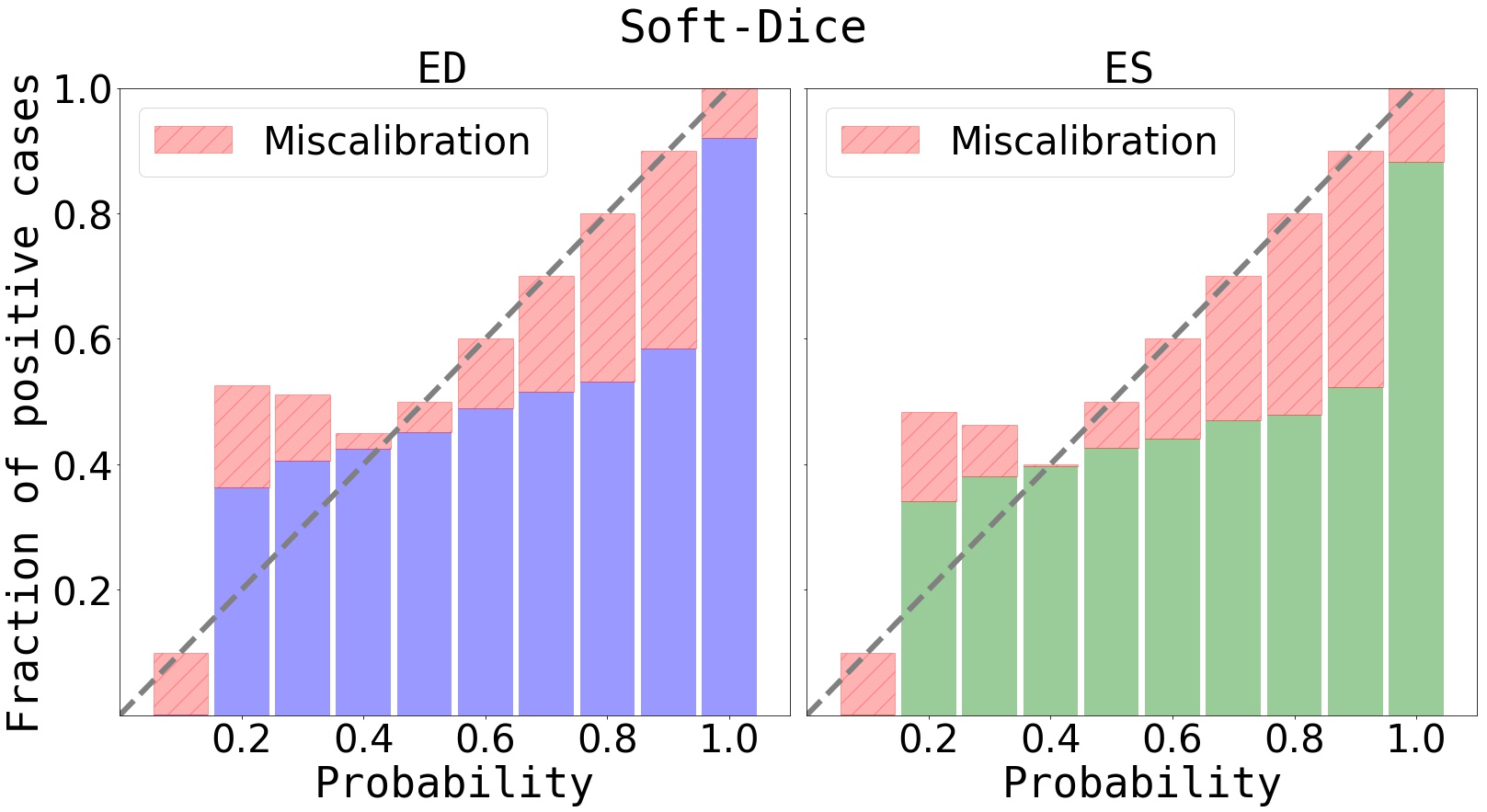}
				\caption{soft-Dice loss}
				\label{fig:calibration_softdice}
				 \end{subfigure}%
			\end{minipage}
			\\[2ex]
			\begin{minipage}{0.40\textwidth}
				\begin{subfigure}{\textwidth}
				\centering
				\includegraphics[width=0.99\textwidth, trim=0 0 0 20mm, clip]{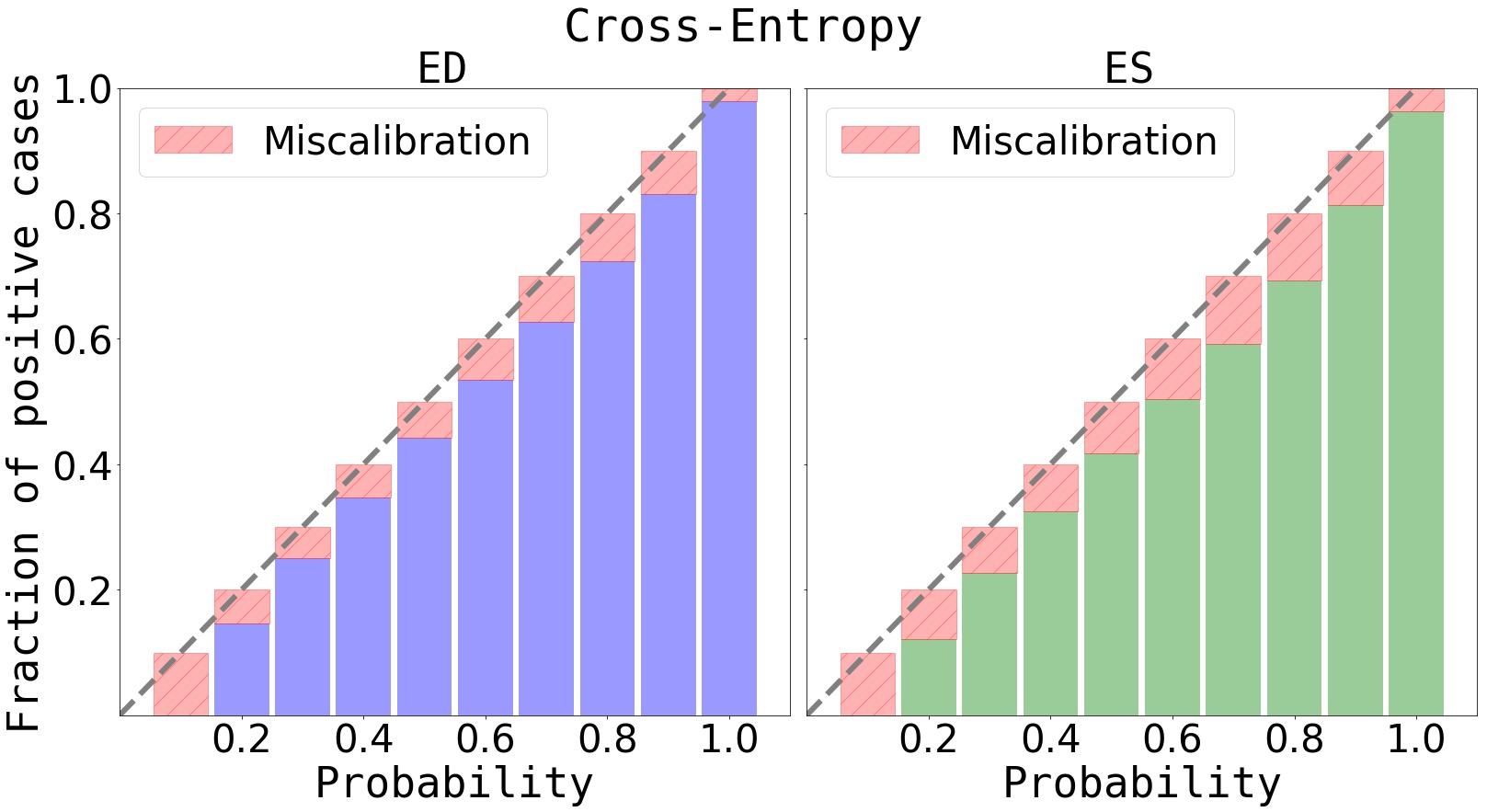}
				\caption{cross-entropy loss}
				\label{fig:calibration_crossent}
				 \end{subfigure}%
			\end{minipage}
			\begin{minipage}{0.40\textwidth}
				\begin{subfigure}{\textwidth}
				\centering
				\includegraphics[width=0.8\textwidth, trim=0 0 0 15mm, clip]{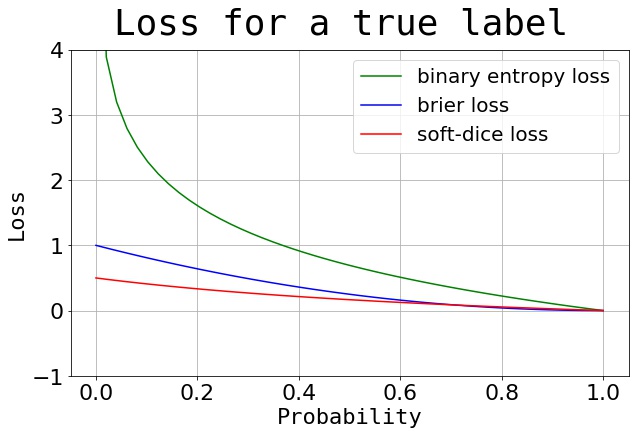}
				\caption{Loss for a true label}
				\label{fig:compare_loss_for_true_positive}
				\end{subfigure}%
			\end{minipage}
			\\[2ex]
			\caption{Reliability diagrams over all tested ED and ES images and tissue classes for Brier, soft-Dice and cross-entropy loss functions. Blue (end-diastole) and green (end-systole) bars quantify the true positive fraction for each probability bin. Red bars quantify the miscalibration of the model where smaller indicates better. If the model is perfectly calibrated, the diagram should match the dashed line.}	\label{fig:reliability_diagram}
	\end{figure}
	 To evaluate whether the obtained per voxel probabilities represent the likelihood of being correct i.e. are well calibrated, we created \textit{Reliability Diagrams} \cite{degroot1983comparison} (RD). Figures.~\ref{fig:calibration_brier}, \ref{fig:calibration_softdice} and \ref{fig:calibration_crossent} show the predicted probabilities discretized into ten bins and plotted against the true positive fraction for each bin. If the model is perfectly calibrated, the diagram should match the dashed line. We conclude that a model trained with the soft-Dice loss produces inferior calibrated probabilities compared to the other two loss functions. We conjecture that this could be caused by the relatively low penalty induced by the soft-Dice loss for the model being \textit{underconfident} for true positive tissue labels (see Fig.~\ref{fig:compare_loss_for_true_positive}). 
   	
   	\begin{figure}
   		\centering
   		\begin{minipage}{0.75\textwidth}
   			\begin{subfigure}{\textwidth}
   				\centering
   				\includegraphics[width=0.99\textwidth]{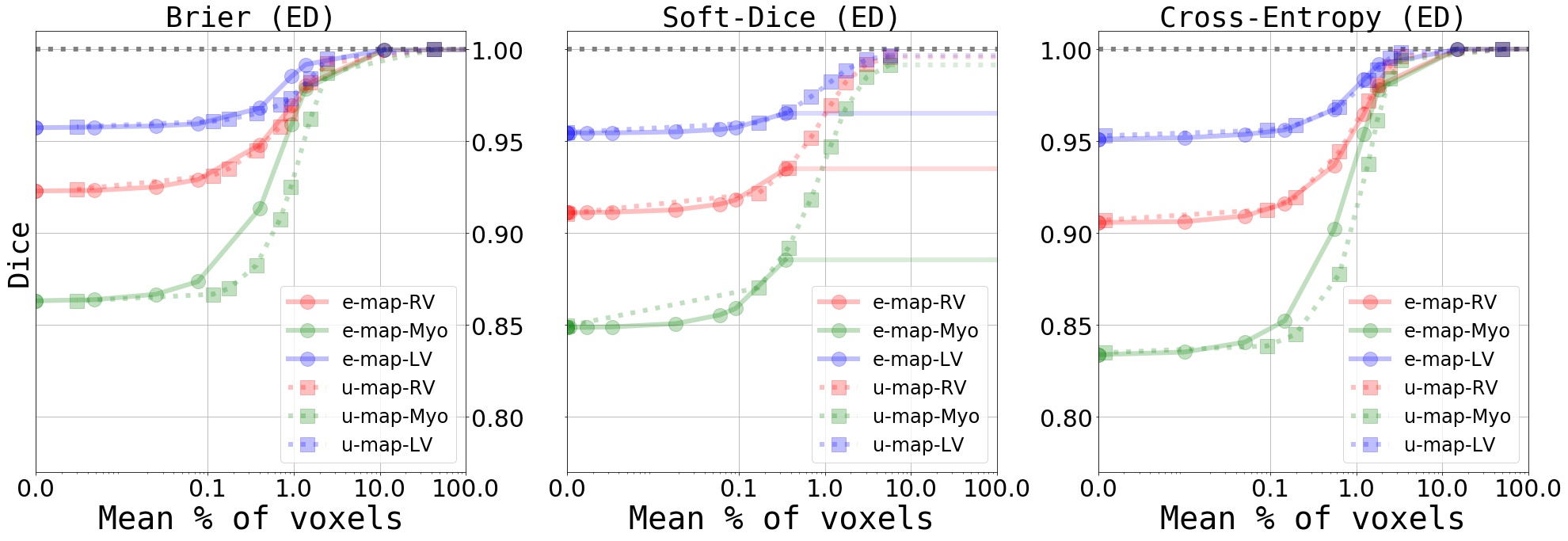}
   				\caption{End-diastole}
   				\label{fig:referral_ed}
   			\end{subfigure}%
   		\end{minipage} 
   		\\[2ex]
   		\centering
   		\begin{minipage}{0.75\textwidth}
   			\begin{subfigure}{\textwidth}
   				\centering
   				\includegraphics[width=0.99\textwidth]{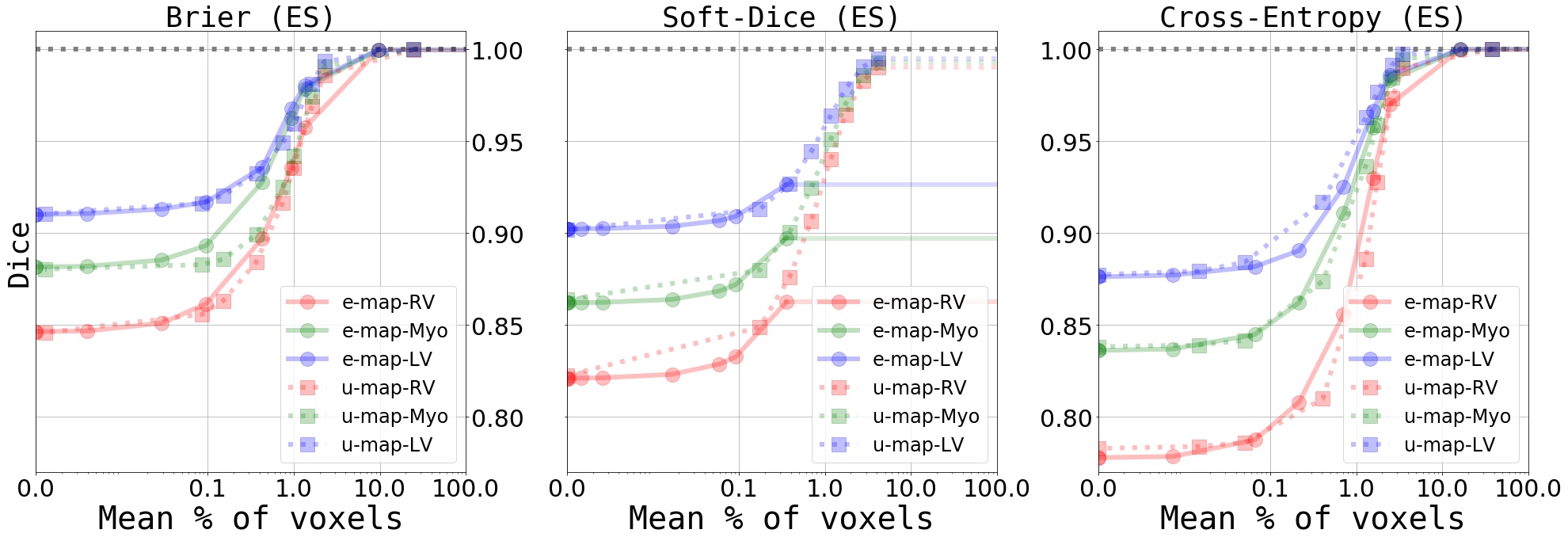}
   				\caption{End-systole}
   				\label{fig:referral_es}
   			\end{subfigure}%
   		\end{minipage}%
   		\\[2ex]
   		\caption{Comparison between entropy and Bayesian uncertainty maps for different loss functions (RV in red, myocardium in green and LV in blue). Figures visualize Dice score of the corrected segmentation mask when voxels above a tolerated uncertainty or entropy threshold are corrected to their reference label. x-axis shows mean percentage of voxels referred in an image.} 
   		\label{fig:uidr_figures}
   	\end{figure}
   	
   	To compare the quality of the obtained uncertainty maps, we simulate a human-in-the-loop setting. We combine the information of predicted segmentation masks with the e-maps or u-maps and assume that voxels above a tolerated uncertainty or entropy threshold are corrected to their reference label by an expert. For each threshold we compute the Dice score for the corrected segmentation mask. Figures~\ref{fig:referral_ed} and \ref{fig:referral_es} visualize the Dice score as a function of the average percentage of voxels thus referred. We observe a monotonic increase in prediction accuracy when more voxels are referred. E.g., inspecting the referral curves for the Brier score loss in Figure~\ref{fig:referral_es} we note that referring on average 1\% of the voxels in an image, increases performance for \num{8}, \num{7} and \num{5}\% for RV, Myo and LV, respectively. These results are similar for the u-maps and the e-maps. In each experiment, the case in which no voxels are referred for correction is considered the baseline (left most y-axis values). We observe that baseline segmentation performance is highest when the model is trained with the Brier score loss, slightly lower for the soft-Dice, and lowest when cross-entropy is used. Except for the soft-Dice loss we note that u-maps and e-maps follow each other quite closely, which suggests that both carry similar information. Not including the soft-Dice loss, segmentation performance with referral using u-maps or e-maps reaches a Dice score of nearly one when sufficient number of voxels are referred. Hence, we may conclude that areas of uncertainty and entropy almost completely cover the regions of incorrect segmentations\footnote{Without covering the complete image in which case all voxels would be referred (corresponding to a trivial solution).}. Results obtained after the referral using entropy maps for a model trained with the soft-Dice loss are clearly inferior compared to the performance achieved when using the u-maps. We assume that this is due to the miscalibration of the model (see Figure~\ref{fig:calibration_softdice}). Compared to e-maps, u-maps tend to exhibit more uncertainty. This is visually expressed for the cross-entropy loss in Figure~\ref{fig:referral_ed}, where the Myo referral-curve obtained with u-maps lags behind the corresponding curve that uses the entropy information. 
   	    \begin{figure}
   		\centering
   		\begin{minipage}{0.19\textwidth}
   			\begin{subfigure}{\textwidth}
   				\centering
   				\includegraphics[width=0.99\textwidth]{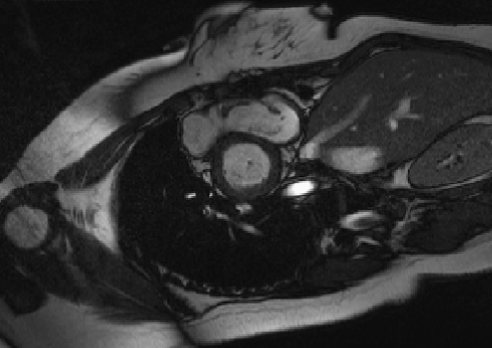}
   				\caption{MRI slice}
   				\label{fig:exp_p100_es_1_mri}
   			\end{subfigure}%
   		\end{minipage}%
   		\centering
   		\begin{minipage}{0.19\textwidth}
   			\begin{subfigure}{\textwidth}
   				\centering
   				\includegraphics[width=0.99\textwidth]{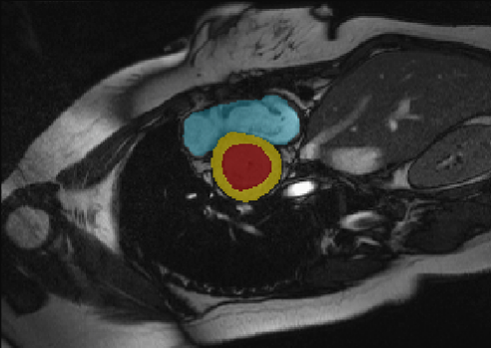}
   				\caption{Reference}
   				\label{fig:exp_p100_es_1_ref}
   			\end{subfigure}%
   		\end{minipage}%
   		\centering
   		\begin{minipage}{0.19\textwidth}
   			\begin{subfigure}{\textwidth}
   				\centering
   				\includegraphics[width=0.99\textwidth]{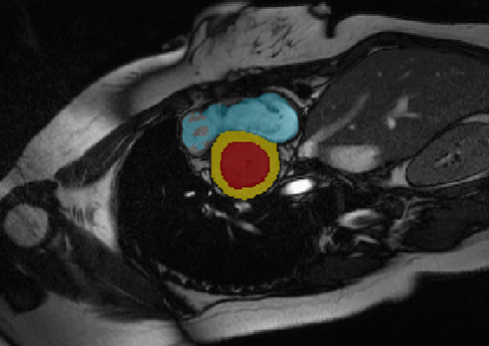}
   				\caption{Automatic}
   				\label{fig:exp_p100_es_1_auto}
   			\end{subfigure}%
   		\end{minipage}
   		\centering
   		\begin{minipage}{0.19\textwidth}
   			\begin{subfigure}{\textwidth}
   				\centering
   				\includegraphics[width=0.99\textwidth]{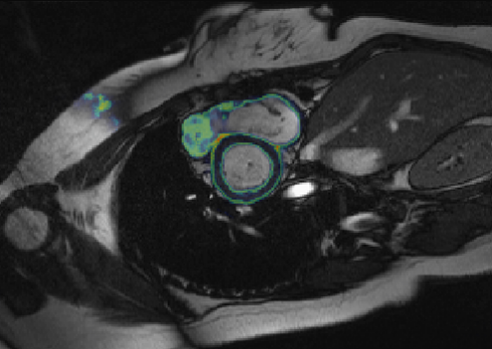}
   				\caption{e-map}
   				\label{fig:exp_p038_ed_4_emap}
   			\end{subfigure}%
   		\end{minipage}%
   		\centering
   		\begin{minipage}{0.19\textwidth}
   			\begin{subfigure}{\textwidth}
   				\centering
   				\includegraphics[width=0.99\textwidth]{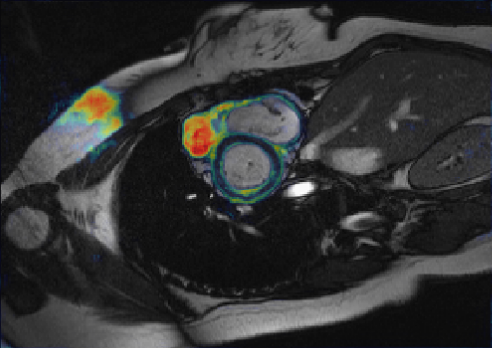}
   				\caption{u-map}
   				\label{fig:exp_p100_es_1_bumap}
   			\end{subfigure}%
   		\end{minipage}%
   		\\[2ex]
   		\caption{Example of segmentation errors that are covered by high uncertainties. Figure~\ref{fig:exp_p100_es_1_mri} shows the original MRI slice to be segmented. Figure~\ref{fig:exp_p100_es_1_ref} visualizes the manual reference segmentation whereas  \ref{fig:exp_p100_es_1_auto} shows the automatic segmentation mask generated by the model when trained with the Brier score loss. Segmentation errors for the right ventricle (in blue \ref{fig:exp_p100_es_1_ref} and \ref{fig:exp_p100_es_1_auto}) are covered by high entropy (figure~\ref{fig:exp_p100_es_1_bumap}) and Bayesian uncertainties (figure~\ref{fig:exp_p038_ed_4_emap}).}
   		\label{fig:exp_brier_bayes_entropy}
   	\end{figure} 
   \begin{figure}
     	\centering
     	\begin{minipage}[t][3cm]{0.2\textwidth}
     		\begin{subfigure}{\textwidth}
     			\centering
     			\includegraphics[width=\textwidth, trim=1.5cm 2cm 3cm .5cm, clip]{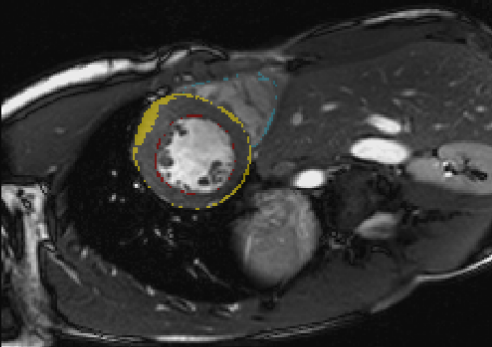}
     			\caption{BS: Errors}
     			\label{fig:exp_all_maps_all_losses_1a}
     		\end{subfigure}%
     	\end{minipage}
     	\begin{minipage}{0.2\textwidth}
     		\begin{subfigure}{\textwidth}
     			\centering
     			\includegraphics[width=\textwidth, trim=1.5cm 2cm 3cm .5cm, clip]{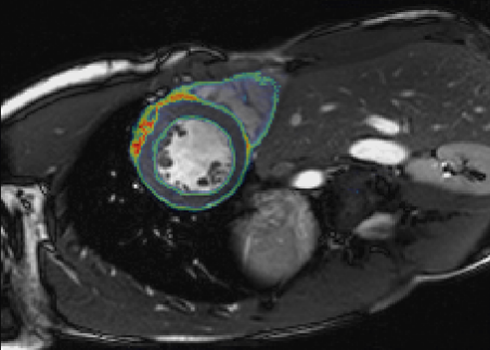}
     			\caption{BS: e-map}
     			\label{fig:exp_all_maps_all_losses_1b}
     		\end{subfigure}%
     	\end{minipage}
     	\begin{minipage}{0.2\textwidth}
     		\begin{subfigure}{\textwidth}
     			\centering
     			\includegraphics[width=\textwidth, trim=1.5cm 2cm 3cm .5cm, clip]{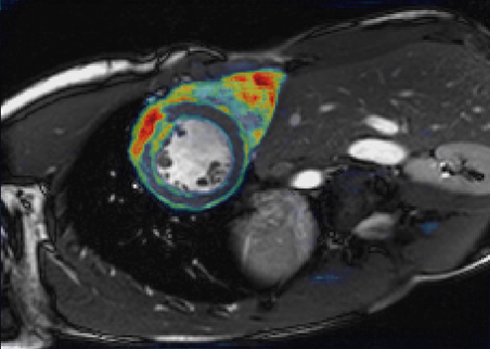}
     			\caption{BS: u-map}
     			\label{fig:exp_all_maps_all_losses_1c}
     		\end{subfigure}%
     	\end{minipage}
        \\
     	\begin{minipage}{0.2\textwidth}
     		\begin{subfigure}{\textwidth}
     			\centering
     			\includegraphics[width=\textwidth, trim=1.5cm 2cm 3cm .5cm, clip]{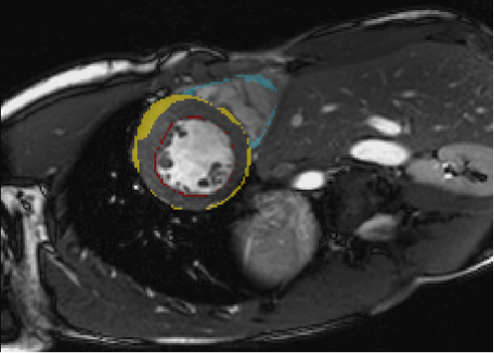}
     			\caption{CE: Errors}
     			\label{fig:exp_all_maps_all_losses_2a}
     		\end{subfigure}%
     	\end{minipage}
     	\begin{minipage}{0.2\textwidth}
     		\begin{subfigure}{\textwidth}
     			\centering
     			\includegraphics[width=\textwidth, trim=1.5cm 2cm 3cm .5cm, clip]{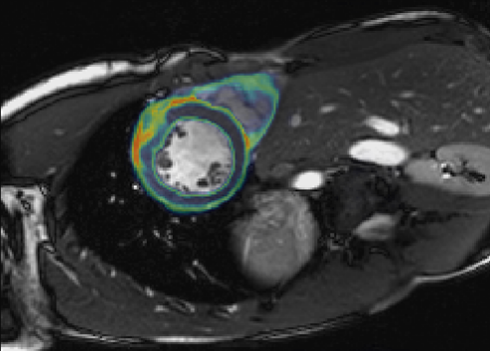}
     			\caption{CE: e-map}
     			\label{fig:exp_all_maps_all_losses_2b}
     		\end{subfigure}%
     	\end{minipage}
     	\begin{minipage}{0.2\textwidth}
     		\begin{subfigure}{\textwidth}
     			\centering
     			\includegraphics[width=\textwidth, trim=1.5cm 2cm 3cm .5cm, clip]{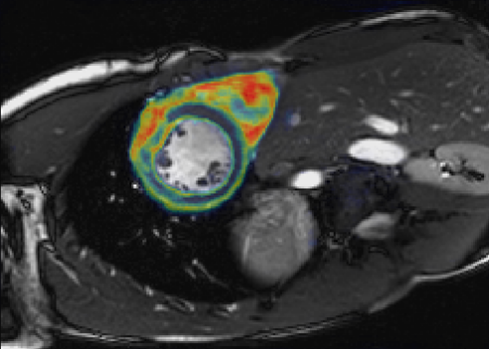}
     			\caption{CE: u-map}
     			\label{fig:exp_all_maps_all_losses_2c}
     		\end{subfigure}%
     	\end{minipage}
     	\\ 
     	\begin{minipage}{0.2\textwidth}
     		\begin{subfigure}{\textwidth}
     			\centering
     			\includegraphics[width=\textwidth, trim=1.5cm 2cm 3cm .5cm, clip]{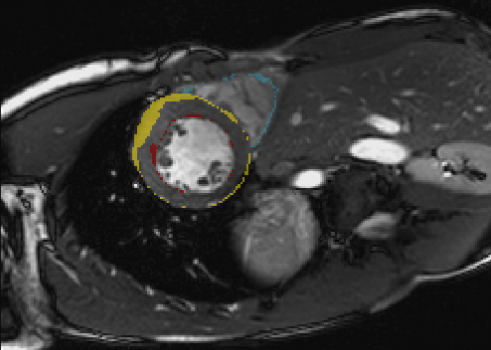}
     			\caption{SD: Errors}
     			\label{fig:exp_all_maps_all_losses_3a}
     		\end{subfigure}%
     	\end{minipage}
     	\begin{minipage}{0.2\textwidth}
     		\begin{subfigure}{\textwidth}
     			\centering
     			\includegraphics[width=\textwidth, trim=1.5cm 2cm 3cm .5cm, clip]{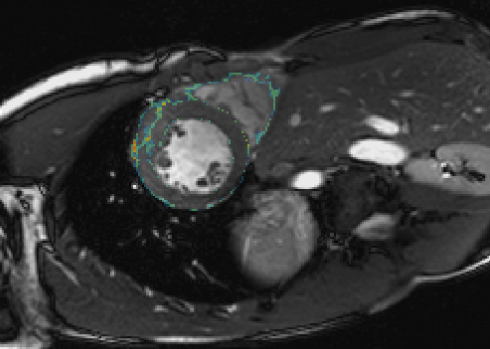}
     			\caption{SD: e-map}
     			\label{fig:exp_all_maps_all_losses_3b}
     		\end{subfigure}%
     	\end{minipage}
     	\begin{minipage}{0.2\textwidth}
     		\begin{subfigure}{\textwidth}
     			\centering
     			\includegraphics[width=\textwidth, trim=1.5cm 2cm 3cm .5cm, clip]{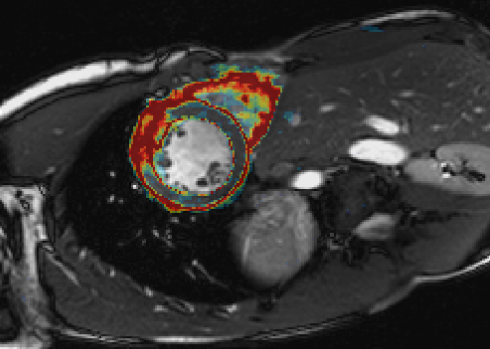}
     			\caption{SD: u-map}
     			\label{fig:exp_all_maps_all_losses_3c}
     		\end{subfigure}%
     	\end{minipage}
     	\\[2ex]
     	\caption{Comparison of (left column) segmentation errors of left ventricle (red), myocardium (yellow) and right ventricle (blue); (middle column) Entropy maps; and (right column) Bayesian uncertainty maps for the Brier score (BS), cross-entropy (CE) and soft-Dice (SD) loss (per row). High uncertainties correspond to red and low uncertainties to blue colors.}
     	\label{fig:exp_all_maps_all_losses}
     \end{figure}
     
   	An example result of the segmentation task performed by a model trained with the Brier score loss is shown in Figure~\ref{fig:exp_brier_bayes_entropy}. The model obviously failed to segment parts of the right ventricle (blue) and we can observe that these errors are covered by entropy and Bayesian uncertainty maps.
   	Figure~\ref{fig:exp_all_maps_all_losses} shows a qualitative comparison of the uncertainty maps for the three different loss functions (corresponding to rows in the figure) that were used during training. Images in the left column visualize the segmentation errors for the three different tissue types using distinct colors. 
   	Although we can observe that the performed errors are roughly the same for the different loss functions, we clearly see significant differences between the uncertainty maps. E.g., when inspecting the e-maps (middle column) we notice that errors with respect to the segmentation of the myocardium are not entirely covered by regions of high uncertainties for a model trained with the soft-Dice loss. In contrast the same regions are almost completely covered by the e-map for a model trained with the Brier score or cross-entropy loss. Furthermore, a model trained with the soft-Dice loss generated u-maps that contain higher uncertainties than u-maps induced by the other two loss functions. We conjecture that this is caused by the miscalibration of the model (see Figure~\ref{fig:calibration_softdice}) which has a bias towards generating probabilities that are close to zero or one, leading to large softmax variances per voxels (we used 10 samples per voxel). This does not affect the e-maps because we do not sample predictions for these maps during testing. Besides, the provided examples corroborate our earlier finding that the u-maps contain more uncertain, yet often correctly segmented voxels than the e-maps.

   	\section{NEW OR BREAKTHROUGH WORK TO BE PRESENTED}
   	This study shows how automatic segmentation can be combined with spatial uncertainty maps to increase the segmentation performance employing a human-in-the-loop setting. Furthermore, our results reveal that we can obtain valuable spatial uncertainty maps with low computational effort using well-calibrated DCNNs.
    \section{Conclusions}
    Using a publicly available cardiac cine MRI dataset, we showed that a (Bayesian) dilated CNN trained with the Brier  loss produces valuable Bayesian uncertainty and entropy maps. Our results convey that regions of high uncertainty almost completely cover areas of incorrect segmentations. Well calibrated models enable us to obtain useful spatial entropy maps, which can be used to increase the segmentation performance of the model.
    
    \acknowledgments
	\noindent This study was performed within the DLMedIA program (P15-26) funded by the Netherlands Organization for Scientific Research (NWO)/ Foundation for Technological Sciences (STW) with contribution by PIE Medical Imaging.

	\bibliography{cardiacMRI}
    \bibliographystyle{spiebib}
    
\end{document}